# Parametric Dependability Analysis through Probabilistic Horn Abduction

Andrea Bobbio, Stefania Montani, Luigi Portinale*
Dipartimento di Informatica
Università del Piemonte Orientale "A. Avogadro"
Alessandria (ITALY)
e-mail: {bobbio,stefania,portinal}@unipmn.it

## Abstract

Dependability modeling and evaluation is aimed at investigating that a system performs its function correctly in time. A usual way to achieve a high reliability is to design redundant systems that contain several replicas of the same subsystem. In order to provide compactness in system representation, parametric system modeling has been investigated in the ,iterature: a set of replicas of a given subsystem is parameterized so that only one representative instance is explicitly included in the model. While modeling aspects can be suitably addressed by these approaches, analytical tools working on parametric characterizations are often more difficult to be defined; the standard approach consists in "unfolding" the parametric model, in order to exploit standard analysis algorithms working at the unfolded "ground" level. In the present paper we consider the formalism of *Parametric Fault Tree* (PFT) and we show how it can be related to *Probabilistic Horn Abduction* (PHA). Since PHA is a framework where both modeling and analysis can be performed in a restricted first-order language, we aim at showing that the conversion of a PFT into a PHA theory allows for an approach to dependability analysis directly exploiting parametric representation. We will show that classical qualitative and quantitative dependability measures can be characterized within PHA; this makes the PHA framework a candidate for PFT analysis, where also posterior probability computation (often neglected in standard Fault Tree analysis) can be naturally performed. A simple example of a multi-processor system with several replicated units is used to illustrate the approach.

## 1 PARAMETRIC FAULT TREE

Parametric Fault Trees (PFT) have been introduced in [2] as a way of efficiently modeling redundant systems for dependability analysis. The basic idea behind PFT stems from the observation that often, due to the redundancy of the system to be modeled, a FT may contain several similar subtrees. To make the description more compact, the similar subtrees may be folded and parameterized, so that only one *representative* is explicitly included in the model; at the same time, the identity of each replica is maintained through a parameter value. A parameter is declared in a node called a *replicator* node or simply *replicator*. Each *replicator* generates as many subtrees as the possible combinations of values of the parameters declared in the node.

Despite its name, an FT is often represented as a DAG (when a basic event is shared by several subtrees). The same holds for a PFT; more formally a PFT is a bipartite DAG whose nodes are either events ($\mathcal{E}$) or gates ($\mathcal{G}$). Besides events and gates, a PFT comprises the following primitive elements: *types* $\mathcal{T}$, *event classes* $\mathcal{EC}$, *parameters* $\mathcal{P}$ and *failure rates* $\mathcal{R}^1$.

Event nodes $\mathcal{E}$ are represented as boxes on the PFT. As in classical FT, there is a single node in $\mathcal{E}$, called the TE (top event), which is not input to any gate node and represents the system failure. Basic events $\mathcal{BE}$ are events for which no further subdivision is necessary (system components) and, therefore, they are not output to any gate node; $\mathcal{BE}$ are denoted by a circle next to the event box.

Gates $\mathcal{G}$ can belong to one of the following categories: AND, OR, implicit ($k : n$); they are denoted using classical gate notation.

$\mathcal{T}$ is a set of finite and disjoint non-empty sets called

---
*Corresponding author

[1]A more formal definition can be found in [2].

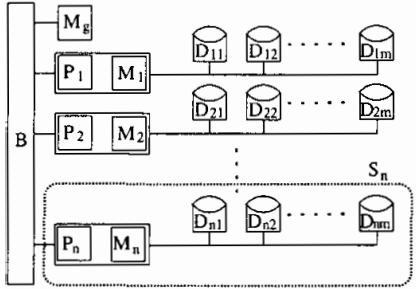

Figure 1: A multiprocessor system.

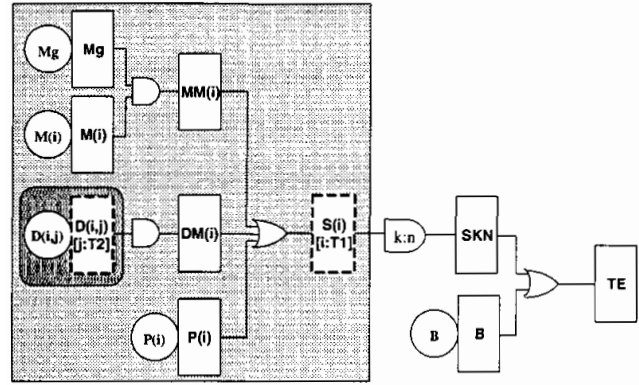

Figure 2: PFT for the system of fig. 1

types.
$\mathcal{EC}$ is the set of event classes, i.e. a collection of similar events. $\mathcal{BEC} \in \mathcal{EC}$ is the subset including only classes of basic events. If $e \in \mathcal{E}$, then $[e]$ denotes the event class to which $e$ belongs.

$\mathcal{P}$ is a set of typed parameters. Parametric events identify a generic event in a given class. A parameter must be associated with a type from $\mathcal{T}$ and $\mathcal{T}_x$ will denote the type of a parameter $x$. A parameter must be declared in exactly one event node. Event nodes where a parameter is declared are called *replicators* ($\mathcal{RE}$) and are represented as dashed boxes.

$\mathcal{R}$ is a function assigning to each basic event class a failure rate, needed for quantitative analysis purposes. If $be \in \mathcal{BE}$, then $\lambda_{[be]}$ will denote the failure rate of the class to which $be$ belongs. In the following, we will assume constant failure rates with an exponential distribution (i.e. the probability of failure of component $be$ at time $t$ is given by $p = 1 - e^{-\lambda_{[be]}t}$).

A constraint imposed by the PFT formalism is that implicit $(k : n)$ gates must have just one input event which is a replicator; this kind of gate is adopted to model voting mechanisms[2].

As an example, consider the multiprocessor system sketched in figure 1 (the example is taken from [2] and it is an extension of another one originally proposed in [7]); it is composed by $n$ independent subsystems $S_1, \ldots, S_n$. Each subsystem $S_i$ ($i = 1, \ldots, n$) is composed by one processor $P_i$ one local memory $M_i$ and $m$ replicated mirrored disk units $D_{i,j}$, where the index $i = 1, \ldots, n$ refers to the subsystem and the index $j = 1, \ldots, m$ numbers the disk replica inside the subsystem. The system redundancy is augmented by a shared common memory $Mg$ that can replace any single local memory $M_i$. A single bus $B$ connects the $n$ subsystems and the shared common memory. The complete system failure occurs when either the bus $B$ fails or with a $(k : n)$ voting mechanism over the $n$ subsystems. Each subsystem fails if the processor fails or all the disks fail or both the local as well as the shared memory fail.

The PFT for the example is drawn in Figure 2 (we emphasize with different gray levels the scope of each parameter). The set of types $\mathcal{T}$ contains two elements: the type $T_1$ of cardinality $n$ identifying the subsystems and the type $T_2$ of cardinality $m$ identifying the disk units inside each subsystem. The event classes $\mathcal{EC}$ correspond to the elementary components (processor $P$, memory $M$, disk $D$, bus $B$) and modules obtained by the combination of elementary components (disk module $DM$, memory module $MM$ and subsystem $S$).

The element of class $P$ (processor) pertaining to subsystem $i \in T_1$ has label $P(i)$, while $D(i,j)$ indicates the event: "failure of disk $j \in T_2$ in subsystem $i \in T_1$". Similarly $M(i)$ indicates the failure of local memory for subsystem $i$. $B$ or $M_g$ are unique events in their classes and do not need to be specified by any parameter. In figure 2, the parameter $i$ of type $T_1$ is declared in the replicator labeled $S(i)$, and the parameter $j$ of type $T_2$ is declared in the replicator labeled $D(i,j)$. Function $\mathcal{R}$ defines the failure rates of disks ($\mathcal{R}(D)$), processors ($\mathcal{R}(P)$), memories ($\mathcal{R}(M)$ and $\mathcal{R}(M_g)$) and the bus ($\mathcal{R}(B)$). Table 1 shows the values adopted in the example.

Table 1: Failure rates for the multiprocessor system model.

| Component | Failure rate (fault/hour) |
|---|---|
| processor | $5 * 10^{-7}$ |
| disk | $8 * 10^{-5}$ |
| local memory | $3 * 10^{-8}$ |
| shared memory | $3 * 10^{-8}$ |
| bus | $2 * 10^{-9}$ |

---

[2]Following dependability conventions, a $(k : n)$ voting means that at least $k$ elements over a total of $n$ has to be working in order to have a (sub)-system working, i.e. $n - k + 1$ elements over $n$ must fail, in order to produce a failure.

The structural complexity of the PFT of Figure 2 does not depend on the number of elements inside each event class. Exactly the same model structure can represent any number of subsystems and any number of disks by suitably setting the cardinality of the corresponding types $T_1$ and $T_2$.

## 2 PFT AND PROBABILISTIC HORN ABDUCTION

Probabilistic Horn Abduction (PHA) has been introduced by Poole as a way of characterizing probabilistic reasoning in terms of the notion of *abductive explanation* [9, 10]. A major feature of the formalim is the use of a first-order Horn language that allows for a compact representation of probabilistic knowledge, through the use of variables and functions as arguments of predicates. In fact, PHA can be viewed as a way of extending an important probabilistic formalism like Bayesian Networks (BN) beyond a propositional language [8, 10].

From the dependability point of view, we have previously shown that any standard FT can be suitably modeled by means of a BN [4, 3]; in particular a BN may be adopted in order to augment both the modeling and the analysis power of a standard FT. However, being a BN a propositional formalism, as in the case of a standard FT, the size of the model may become considerable when modeling a redundant system with several replicas. In our multiprocessor example, any additional replica of a subsystem $i \in T_1$ implies that $m + 5$ ($m \in T_2$) more event nodes (and 3 more gates) have to be added to a standard FT (i.e. $m + 5$ more nodes have to added to the corresponding BN). It follows that in a highly redundant system, the number of nodes (in either a FT or a BN) can become quite large. A PFT has no such problems, since the structure complexity is independent from the number of replicas.

The aim of the paper is to show that it is possible to convert a PFT into a corresponding PHA theory where the parametric representation can be (at least partially) preserved; in addition, PHA-based algorithms such as best-first top-down search of explanations [9] can be fruitfully adopted in order to compute dependability measures. Qualitative measures such as *minimal cut-sets* (MCS) or quantitative measures (such as system unreliability, posterior probability of faults, etc...) can be computed within the PHA framework. This provides the PFT formalism with an effective analysis methodology.

First of all, let us briefly remind the basics of PHA (see [10] for more details).

A **disjoint declaration** is of the form

$$disjoint([h_1 : p_1, \ldots h_n : p_n])$$

where $h_i$ are atoms (called *hypotheses*), $p_i$ are real numbers $0 \leq p_i \leq 1$ such that $\sum_i p_i = 1$. Any variable appearing in one $h_i$ must appear in all of the $h_j$ and the declaration implies the clause $false \leftarrow h_i \wedge h_j$ for $i \neq j$. Number $p_i$ denotes the (prior) probability of $h_i$.

A **Probabilistic Horn Abduction Theory** is a collection of definite clauses and disjoint declarations such that a ground atom $h$ can be instance of a hypothesis only in one disjoint declaration.

An **Abductive scheme** is a pair $\langle F, H \rangle$ where $F$ is a set of Horn clauses and $H$ a set of atoms; let $H'$ be the set of ground instances of elements of $H$. If $g$ is a closed formula, an *explanation* of $g$ from scheme $\langle F, H \rangle$ is $D \subseteq H'$ such that $F \cup D \models g$ and $F \cup D \not\models false$. A *minimal explanation* of $g$ is an explanation of $g$ such that no strict subset is an explanation of $g$.

## 3 CONVERTING A PFT INTO A PHA THEORY

While converting a PFT into PHA, several aspects have to be taken into account: the presence in the PFT of different kinds of events (basic events, replicators, etc...), the parametric form of the events, the logical behavior of gates and the probabilistic characterization of the involved events. Concerning the last point, it is worth remembering that in a PHA theory a set of assumptions have to be satisfied, in order to perform correct probability computations or estimates. We will return on this point in the following; for now, we first provide a conversion from PFT to PHA as direct as possible and we show what kind of analysis can be performed at that level of conversion. Subsequently, we will show how this conversion has to be refined, in order to have the PHA theory fulfilling basic assumptions leading to a consistent probability analysis in terms of abductive explanations.

Let $P$ be a PFT with events $\mathcal{E}$ (basic events $\mathcal{BE} \subseteq \mathcal{E}$, replicator events $\mathcal{RE} \subseteq \mathcal{E}$), event classes $\mathcal{EC}$, gates $\mathcal{G}$, types $\mathcal{T}$, parameters $\mathcal{P}$ and event class failure rates $\mathcal{R}$. Let $t$ be the analysis time of the PFT; a PHA theory $F(t)$ corresponding to $P$ at time $t$ is generated as follows:

**Conversion 1.**

1. for each basic event $be(i_1, \ldots i_n) \in \mathcal{BE}$ and for each n-uple $(v_1, \ldots v_n)$ with $v_j \in \mathcal{T}_{i_j}$, create a disjoint declaration like

   $$disjoint([be(v_1, \ldots v_n, f) : p, be(v_1, \ldots v_n, w) : 1-p])$$

68 BOBBIO ET AL. UAI 2003

where arguments $f$ and $w$ of predicate $be$ stand for *failed* and *working* behavior of the corresponding component respectively and $p = 1 - e^{-\lambda_{[be]}t}$

2. for each gate $g \in \mathcal{G}$, let $e(u_1, \ldots u_n) \in \mathcal{E}$ be the output event of $g$ and $e_i(x_1 \ldots x_{n_i})$ $1 \leq i \leq m$ the set of $m$ input events for $g$;

   **2.a** if $g$ is an OR gate, create $m$ clauses

   $$e(u_1, \ldots u_n) \leftarrow e_i(x_1, \ldots x_{n_i}) \text{ if } e_i \notin \mathcal{BE}$$

   $$e(u_1, \ldots u_n) \leftarrow e_i(x_1, \ldots x_{n_i}, f) \text{ if } e_i \in \mathcal{BE}$$

   **2.b** if $g$ is an AND gate, create a clause

   $$e(u_1 \ldots u_n) \leftarrow f_1 \wedge f_2 \ldots \wedge f_m$$

   where

   $$f_i \equiv \begin{cases} e_i(x_1, \ldots x_{n_i}) & \text{if } e_i \notin \mathcal{RE}, e_i \notin \mathcal{BE} \\ e_i(x_1, \ldots x_{n_i}, f) & \text{if } e_i \notin \mathcal{RE}, e_i \in \mathcal{BE} \end{cases}$$

   $$f_i \equiv \begin{cases} \bigwedge_{(z_1, \ldots z_{n_i})} e_i(z_1 \ldots z_{n_i}) & \text{if } e_i \in \mathcal{RE}, e_i \notin \mathcal{BE} \\ \bigwedge_{(z_1, \ldots z_{n_i})} e_i(z_1 \ldots z_{n_i}, f) & \text{if } e_i \in \mathcal{RE}, e_i \in \mathcal{BE} \end{cases}$$

   with

   $$z_j = \begin{cases} x_j & \text{if } x_j \text{ is not declared in } e_i \\ v_j \in \mathcal{T}_{x_j} & \text{if } x_j \text{ is declared in } e_i \end{cases}$$

If a gate $g$ is of type implicit $(k : n)$, it is trivial to verify that it can be modeled via a set of AND/OR gates, so that there is no need to give a specific translation for it (see [2]). While the conversion of an OR gate is quite immediate to understand, the case of the AND gate deserves some discussion: indeed, if a replicator is input to an AND gate, the semantics of the PFT implies that the conjunction of the set of events generated from the replicator (one for each possible instantiation of the set of parameters declared in the replicator) has to be considered. This is why, if the input event $e_i$ is a replicator, the body of the clause is the conjunction of all the predicates corresponding to events generated by each possible instantiation of the parameters declared in the replicator.

**Example.** Consider the PFT of fig. 2 of the multiprocessor system of fig. 1; suppose that the number of disks in each subsystem is $m = 2$, the number of replicated subsystems is $n = 3$ and that the voting mechanism is $(2 : 3)$ ($k = 2, n = 3$). Suppose also that type $T_1 = \{1, 2, 3\}$ (subsystem identities) and that type $T_2 = \{1, 2\}$ (disk identities within a subsytem). By considering an analysis time $t = 10,000$ hours, we get the following PHA theory (we use a Prolog-like notation for convenience where ":-" stands for $\leftarrow$, "," for conjunction and upper case letters denotes variables).

```
/* Local Memories */
disjoint([m(1,w):0.9997,m(1,f):0.0003])
disjoint([m(2,w):0.9997,m(2,f):0.0003])
disjoint([m(3,w):0.9997,m(3,f):0.0003])

/* Global Memory */
disjoint([mg(w):0.9997,mg(f):0.0003])

/* Local Disks */
disjoint([d(1,1,w):0.4493,d(1,1,f):0.5507])
disjoint([d(1,2,w):0.4493,d(1,2,f):0.5507])
disjoint([d(2,1,w):0.4493,d(2,1,f):0.5507])
disjoint([d(2,2,w):0.4493,d(2,2,f):0.5507])
disjoint([d(3,1,w):0.4493,d(3,1,f):0.5507])
disjoint([d(3,2,w):0.4493,d(3,2,f):0.5507])

/* Processors */
disjoint([p(1,w):0.9950,p(1,f):0.0050])
disjoint([p(2,w):0.9950,p(2,f):0.0050])
disjoint([p(3,w):0.9950,p(3,f):0.0050])

/* Global Bus */
disjoint([b(w):0.99998,b(f):0.00002])

mm(I):- mg(f),m(I,f)

dm(I):- d(I,1,f),d(I,2,f)

s(I):- mm(I)
s(I):- dm(I)
s(I):- p(I,f)

skn:- s(1),s(2)
skn:- s(1),s(3)
skn:- s(2),s(3)

te:- b(f)
te:- skn
```

We can notice that, apart from the *disjoint* declarations (where each single component of the system has to be introduced with its identity) the clauses of the theory are almost completely parametric. Parameter instatiation becomes necessary when dealing with replicators involved as input of AND gates (and consequently in the case of replicators input to implicit $(k : n)$ gates). In the above example, event $D(i, j)$ is a replicator for the declared parameter $j \in T_2$, input to an AND gate with event $DM(i)$ as output. Atom d(I,J,f) is used to model the event $D(i, j)$ (we use the 'f' argument to model the failure, since $D(i, j)$ is a basic event) and atom dm(I) is introduced to model the parametric event $DM(i)$. The AND gate is then modeled through the clause dm(I):-d(I,1,f),d(I,2,f) where the body is the conjunction of atoms d(I,J,f) for each possible value of variable $J$, corresponding to the declared parameter $j \in T_2$ (i.e. for each possible value of $T_2 = \{1, 2\}$). Notice that atoms d(I,1,f) and d(I,2,f) are still parametric on $I$, since parameter $i$, corresponding to PHA variable $I$, is not declared inside the replicator $D(i,j)$. The other replicator in the example is event $S(i)$ input to the (2 : 3) gate and declaring $i$; this is modeled by means of the three clauses having atom skn (corresponding to event $SKN$ in the PFT) in the head.

## 4 DETERMINING MINIMAL CUT-SETS

In dependability analysis, the most important qualitative measure obtainable fro a FT concerns so called *Minimal Cut-Sets* (MCS); each MCS is a set of basic events that is a prime implicant of the TE. MCS correspond to the minimal sets of basic components of the modeled system that can be considered responsible of a fault.

The **Conversion 1** procedure illustrated in section 3 is just a re-writing of the gates of a PFT in terms of definite clauses, where parameters of the PFT becomes variables in the clauses. Basic events are listed in *disjoint declarations* and they form the set of hypotheses that can be used to explain a given atom in the corresponding PHA theory. In particular, by explaining the atom corresponding to the TE, we can build abductive explanations that represent (not necessarily minimal) implicants (i.e. cut-sets of the TE).

In [9], Poole presented a top-down best-first abductive procedure for building explanations of a given atom whithin PHA[3]; such a procedure represents an anytime algorithm able to provide one explanation at the time, in order of probability. It can then be adopted in order to compute generic cut-sets of the TE, by explaining the te atom corresponding to TE. Unfortunately, minimality of explanations cannot be guaranteed a priori; on the other hand, Poole's algorithm is best-first, using prior probability of explanations as evaluation function: a given explanation (cut-set) is then generated before any other explanation having lower probability. Basic events in a PFT are assumed to be mutually independent and this is modeled by the fact that in PHA the same assumption holds for atoms declared in different disjoint declarations; this means that a non-minimal cut-set (i.e. a non-minimal explanation) has certainly lower probability than each MCS that it contains. In algorithmic terms, this implies that each MCS is always generated before any non-minimal cut-set containing it. A simple minimality check can then be performed every time a new explanation is generated: if there is an already generated explanation which is contained in the current one, then the latter is discarded. Using this simple approach we are able to generate the MCS of the given PFT, by simply explaining the te atom in the corresponding PHA theory provided by **Conversion 1**.

---

[3]For the experiments described in the present paper we used the code downloaded from http://www.cs.ubc.ca/spider/poole/code.html.

Table 2: Top 13 MCS and their (prior) unreliability at $t = 10^4$ hours

| MCS | Prior Prob. |
|---|---|
| $\{D(1,1), D(1,2), D(2,1), D(2,2)\}$ | 0.091954 |
| $\{D(1,1), D(1,2), D(3,1), D(3,2)\}$ | 0.091954 |
| $\{D(2,1), D(2,2), D(3,1), D(3,2)\}$ | 0.091954 |
| $\{D(1,1), D(1,2), P(2)\}$ | 0.001512 |
| $\{D(1,1), D(1,2), P(3)\}$ | 0.001512 |
| $\{D(2,1), D(2,2), P(3)\}$ | 0.001512 |
| $\{D(2,1), D(2,2), P(1)\}$ | 0.001512 |
| $\{D(3,1), D(3,2), P(2)\}$ | 0.001512 |
| $\{D(3,1), D(3,2), P(1)\}$ | 0.001512 |
| $\{P(1), P(2)\}$ | 0.000025 |
| $\{P(1), P(3)\}$ | 0.000025 |
| $\{P(2), P(3)\}$ | 0.000025 |
| $\{B\}$ | 0.00000003 |

In our multiprocessor example, by running Poole's top-down search with minimality check, we obtain 28 MCS. Table 2 reports the top 13 MCS ranked by their prior probability computed at time $t = 10^4$ hours. This represents a quantitative measure used in standard FT analysis, called the *MCS unreliability*; as we will see in the next section, computing posterior unreliability (given that a fault has occurred) can be more informative, in order to estimate a precise criticality of components. From table 2, we can just notice that a disk failure in two subsystems is more probable than a disk failure in one subsystem together with a processor failure in another subsystem, which is more critical than a processor failure in two subsystems, which is more critical than a bus failure. However, there is no precise "weighting" of such criticality, given that the system fault has occurred (i.e. given that TE is true).

It is worth noting that the MCS obtained are not parametric, since the hypotheses used in the disjoint declarations are ground atoms. As noticed in [2], having parametric cut-sets would reduce the number of elements to be considered, since a parametric cut-set is a representative of a set of cut-sets; however, the computation of parametric MCS for a PFT is currently still an open problem; for example, methods based on the translation of the PFT into a Petri net (as proposed in [2] provide non-parametric cut-sets like the approach described above.

## 5 QUANTITATIVE ANALYSIS

In the section 4 we discussed how **Conversion 1** can be used for qualitative analysis, by exploiting top-down best-first search on the PHA theory in order to compute *minimal explanations* of the TE, i.e. MCS. However, no quantitative analysis can be performed on the resulting theory. Indeed, in [10] it is shown that, in order to correctly compute (or estimate) posterior probabilities of atoms, two basic assumptions have to be satisfied by a PHA theory: *(1) there are no clauses in the theory whose head unifies with a hypothesis; (2) if $F'$ is the set of ground instances of the elements of the PHA theory $F$, the bodies of the clauses in $F'$ are mutually exclusive.* Under both assumptions, minimal explanations of conjunctions of atoms are mutually exclusive; this means that given a conjunction $g$, if $expl(g, F)$ is the set of minimal explanations of $g$ in the PHA theory $F$, the probability of $g$ can be computed as

$$P(g) = \sum_{e_i \in expl(g,F)} P(e_i). \quad (1)$$

While the first assumption is satisfied by **Conversion 1**, the second one is not; for instance, in case of an OR gate, input events are independently modeled as predicates in the bodies of separate clauses, without any guarantee of mutual exclusion.

Fortunately, the restriction of assumption (2) is just syntactic, since it is possible to rewrite the knowledge base in such a way that assumption (2) is satisfied (see [10] for the details).

**Conversion 2.**

1. use procedure **Conversion 1** to generate a PHA theory $F_1$;

2. transform each set of clauses in $F_1$ having the same head into a set of clauses having disjoint bodies, following the procedure described in [10] and producing a PHA theory $F_2$.

This process implies the construction of atoms representing the negation of other atoms; for this reason we have to explicitly introduce the working ($w$) and failure status ($f$) also for events which are not basic events. Using this approach, the PHA theory introduced in section 3 transforms in the following one (we omit disjoint declarations that remain unchanged):

```
mm(I,f):- mg(f),m(I,f)
mm(I,w):- mg(w)
mm(I,w):- m(I,w),mg(f)

dm(I,f):- d(I,1,f),d(I,2,f)
dm(I,w):- d(I,1,w)
dm(I,w):- d(I,2,w),d(I,1,f)

s(I,f):- p(I,f)
s(I,f):- mm(I,f),p(I,w)
s(I,f):- dm(I,f),mm(I,w),p(I,w)
s(I,w):- dm(I,w),mm(I,w),p(I,w)

skn(f):- s(1,f),s(2,f)
skn(f):- s(1,f),s(3,f),s(2,w)
skn(f):- s(2,f),s(3,f),s(1,w)
skn(w):- s(1,w),s(2,w)
skn(w):- s(1,w),s(3,w),s(2,f)
skn(w):- s(2,w),s(3,w),s(1,f)

te:- b(f)
te:- skn(f),b(w)
```

As noticed in [10], this approach biases the most probable explanations to less specific clauses; however, we do not aim to use the above model for qualitative purposes, but for quantitative analysis only. In particular on the above model, we can exploit top-down best-first search to compute arbitrary conditional probability. In dependability terms, the most important quantitative measure is the *system unreliability* at time $t$. This represents the probability of failure in the system at time $t$. Let $F_2(t)$ be the PHA theory produced by **Conversion 2** at analysis time $t$; system unreliability reduces to compute the probability of the TE, by explaining the corresponding atom te in $F_2(t)$:

$$P(TE) = \sum_{e_i \in expl(\text{te}, F_2(t))} P(e_i)$$

Changing analysis time simply corresponds to change probability values in disjoint declarations. Figure 3 plots the system unreliability of our multiprocessor example over time, computed with Poole's algorithm, between 0 and 20000 hours, with a step of 2000 hours[4]. As discussed in section 4, another important quantitative measure is the MCS *unreliability*, usually represented by the joint (prior) probability of the events corresponding to the cut-set: it is computed by assuming the independence of the failure events of system components. As show in [3], considering posterior probabilities of events given that a system failure has occurred, can provide a more reliable analysis of the criticality of system components; in [3] we showed that, in case of a standard FT, the use of BN posterior probability computation can be suitably adopted for this aspect. In case of a PFT model, the corresponding PHA theory $F_2(t)$ obtained by **Conversion 2** at time $t$ can be used in a similar way. If $C$ is a MCS and c is the conjunction of atoms corresponding to $C$ in the PHA theory, because of equation (1), posterior

---

[4]These results have been verified by performing the same computation on the BN corresponding to the unfolded versions (at different times) of the PFT of fig. 2, through the JavaBayes tool [5]
(http://www.pmr.poli.usp.br/ltd/Software/javabayes).

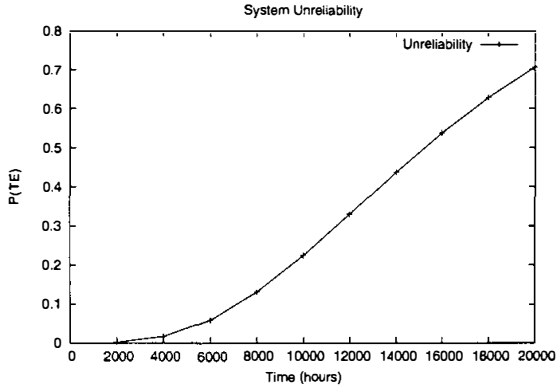

Figure 3: System Unreliability over Time

Table 3: Top 13 MCS and their posterior unreliability at $t = 10^4$ hours

| MCS | Post. Unreliab. |
|---|---|
| $\{D(1,1), D(1,2), D(2,1), D(2,2)\}$ | 0.409541 |
| $\{D(1,1), D(1,2), D(3,1), D(3,2)\}$ | 0.409541 |
| $\{D(2,1), D(2,2), D(3,1), D(3,2)\}$ | 0.409541 |
| $\{D(1,1), D(1,2), P(2)\}$ | 0.006736 |
| $\{D(1,1), D(1,2), P(3)\}$ | 0.006736 |
| $\{D(2,1), D(2,2), P(3)\}$ | 0.006736 |
| $\{D(2,1), D(2,2), P(1)\}$ | 0.006736 |
| $\{D(3,1), D(3,2), P(2)\}$ | 0.006736 |
| $\{D(3,1), D(3,2), P(1)\}$ | 0.006736 |
| $\{P(1), P(2)\}$ | 0.000111 |
| $\{P(1), P(3)\}$ | 0.000111 |
| $\{P(2), P(3)\}$ | 0.000111 |
| $\{B\}$ | 0.000089 |

unreliability of $C$ can be computed as

$$P(C|TE) = \frac{\sum_{e_i \in expl((c \wedge te), F_2(t))} P(e_i)}{\sum_{e_j \in expl(te, F_2(t))} P(e_j)}$$

Table 3 shows the top 13 MCS of the example with their posterior unreliability at $t = 10^4$ hours. We can notice that, while the ranking of MCS does not change with respect to table 2, now the quantitative measure computed (i.e the posterior probability of each MCS given the TE) is an effective weight of their criticality with respect to the system fault. For instance, we can now state that there is about 40% of probability that, in case of system fault, disk failures in two subsytems are the responsible for the fault. Notice that, since MCS are implicants of the TE, the conditional probability of TE given a MCS is 1, so the posterior of a MCS given the TE differs from the prior only for the constant $P(TE)^{-1}$. Even if such values are in principle computable also with FT analysis (after the computation of the probability of the TE), FT-based tools usually just report the information of table $2^5$.

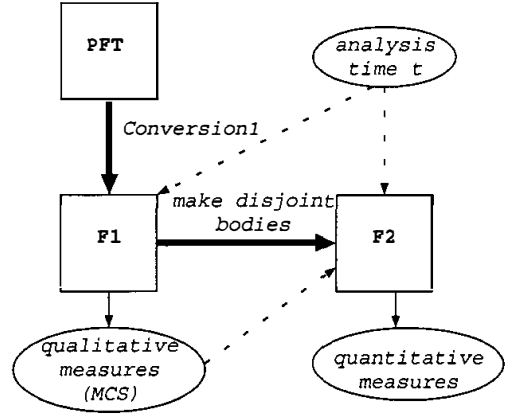

Figure 4: Using PHA for PFT analysis

Figure 4 epitomizes the whole approach: from a given PFT, by means of **Conversion 1** a PHA theory $F_1$ is generated from which MCS can be obtained; from $F_1$, by making bodies of clauses disjoint, we get a new theory $F_2$ from which quantitative analysis can be performed. **Conversion 2** is then represented by the sequential application of the steps corresponding to the two bold arrows. Dashed arrows represent possible inputs. In particular the analysis time $t$ is fundamental for quantitative analysis using $F_2$, but is necessary to $F_1$ as well; indeed, since best-first search of MCS on $F_1$ uses prior probabilities of hypotheses, we need to compute a value of such probabilities using a given analysis time. MCS may then be used as input to quantitative analysis (for example for computing their posterior unreliability).

Of course, any kind of posterior probability computation can be performed on $F_2$, by making possible different kind of diagnostic inferences usually neglected in classical dependability analysis. For instance, we can analyse the criticality of each single component by computing the posterior probability of each basic event given the TE. Table 4 reports the results obtained by running the posterior computation algorithm at time $t = 10^4$ hours. Notice that, differently from posterior unreliability of MCS, these values cannot be computed by just normalizing prior probabilities using TE probability.

## 6　CONCLUSION

Parametric modeling of redundant systems is fundamental in dependability analysis, since it provides a compact system representation; PFT have been proposed as a way to achieve such a compactness. In the present paper we have related PFT to the PHA formalism, where the parametric representation can be partially preserved. The use of search algorithms on a PHA theory can be effectively used to compute reliability measures, both qualitative and quantitative; this provides an analysis approach to PFT at the parametric level.

---
[5]Some recent works on FT analysis address the importance of posterior probability computation as well (see [6]).

Table 4: . Posterior probabilities of basic events at time $t = 10^4$ hours ($i \in T_1, j \in T_2$).

| Basic Event | Posterior Prob. |
|---|---|
| $D(i,j)$ | 0.8074582 |
| $P(i)$ | 0.0115368 |
| $M(i)$ | 3.001e-4 |
| $Mg$ | 3.003e-4 |
| $B$ | 8.91e-5 |

Since PHA is a generalization of BN to a first-order language, several advantages can be obtained with respect to standard FT analysis (as also discussed in [3, 4]), in particular with respect to posterior probability computation. Some other researchers have previously investigated the use of BN formalisms for reliability [1, 12, 11], however they always consider non parametric models.

Another advantage of PHA is that top-down search can be performed in any-time fashion [9]; while for the example used in this paper we always computed exact values, for more complex systems it may be more reasonable to just estimate reliability measures. Poole's algorithm can be stopped at any time and the current probability estimate provided as output; for example we can stop the computation of the system unreliability after a given number of explanations of the TE have been computed or when the estimation error is below a given threshold.

Finally, we did not discuss here some modeling advantages that may be provided by PHA with respect to plain PFT that are of the same nature of those obtainable from a BN with respect to a simple FT: they range from noisy gates, to multi-state variables, to sequentially dependent faults (see [3, 4] for more details). Our future works will concentrate on defining a principled way with which such aspects can be fruitfully exploited whithin PHA.

**Acknowledgements**

This work has been partially funded by MIUR under FIRB grant n. RBNE019N8N.